\documentclass[conference]{IEEEtran}
\IEEEoverridecommandlockouts
\usepackage{cite}
\usepackage{amsmath,amssymb,amsfonts}
\usepackage{algorithmic}
\usepackage{graphicx}
\usepackage{textcomp}
\usepackage{xcolor}
\usepackage{multirow}
\usepackage{hyperref}
\usepackage{booktabs}

\def\BibTeX{{\rm B\kern-.05em{\sc i\kern-.025em b}\kern-.08em
    T\kern-.1667em\lower.7ex\hbox{E}\kern-.125emX}}

\begin{document}

\title{Deep Perspective Transformation Based Vehicle Localization on Bird's Eye View}

\author{\IEEEauthorblockN{1\textsuperscript{st} Abtin Mahyar}
\IEEEauthorblockA{\footnotesize{Department of Computer and Data Sciences,} \\
\textit{Shahid Beheshti University,}\\
Tehran, Iran \\
a.mahyar@mail.sbu.ac.ir}
\and
\IEEEauthorblockN{2\textsuperscript{nd} Hossein Motamednia}
\IEEEauthorblockA{\footnotesize{High Performance Computing Laboratory,} \\
\textit{School of Computer Science,} \\
\footnotesize{Institute for Research in Fundamental Sciences,}\\
Tehran, Iran \\
h.motamednia@ipm.ir}
\and
\IEEEauthorblockN{2\textsuperscript{nd} Dara Rahmati}
\IEEEauthorblockA{\footnotesize{Faculty of Computer Science and engineering,} \\
\textit{Shahid Beheshti University,}\\
Tehran, Iran \\
d\_rahmati@sbu.ac.ir}
}



\maketitle
\begin{abstract}

An accurate understanding of a self-driving vehicle's surrounding environment is crucial for its navigation system. To enhance the effectiveness of existing algorithms and facilitate further research, it is essential to provide comprehensive data to the routing system. Traditional approaches rely on installing multiple sensors to simulate the environment, leading to high costs and complexity. In this paper, we propose an alternative solution by generating a top-down representation of the scene, enabling the extraction of distances and directions of other cars relative to the ego vehicle. We introduce a new synthesized dataset that offers extensive information about the ego vehicle and its environment in each frame, providing valuable resources for similar downstream tasks. Additionally, we present an architecture that transforms perspective view RGB images into bird's-eye-view maps with segmented surrounding vehicles. This approach offers an efficient and cost-effective method for capturing crucial environmental information for self-driving cars. \footnote{Code and dataset are available at \href{https://github.com/IPM-HPC/Perspective-BEV-Transformer}{https://github.com/IPM-HPC/Perspective-BEV-Transformer}.}


\end{abstract}


\section{Introduction}

Self-driving vehicles rely on multiple sensors, such as Lidar, Radar, and distance sensors, to obtain a comprehensive representation of the surrounding environment \cite{saoudi2022autonomous}. RGB cameras are also commonly used to detect traffic signs, traffic lights, and street lanes. These cameras offer a cost-effective alternative to other sensors for extracting features for navigation algorithms \cite{badue2021self}. However, the use of multiple sensors in this field presents challenges due to the high cost and complexity involved. Recent studies have attempted to extract a similar representation of the environment using only RGB cameras. To extract characteristics from RGB photos, both Deep Learning (DL) and traditional image processing techniques have been applied.

One valuable piece of information extracted from RGB images is the detection of object locations and distances between vehicles in the captured images and the ego vehicle. Additionally, the direction of surrounding vehicles can be determined, enabling predictions of their future movements \cite{yao2019egocentric}. These types of information can be effectively aggregated into a single top-down representation, also called as a bird's-eye-view (BEV) image, which provides valuable insights for assessing the navigation system of self-driving vehicles \cite{ma2022vision}.

Classical image processing methods have utilized homography matrices to transform RGB images into a 2D top-down representation \cite{mallot1991inverse}. These methods involve calculating the transfer matrix values by adjusting the camera, allowing each pixel of the RGB image to be mapped to the top-down representation. However, one challenge of these methods lies in the need to update the transfer matrix values when the camera's location is changed to ensure proper transformation. DL approaches have also been employed for this task, achieving higher accuracy and efficiency compared to classical methods. Deep neural networks have the advantage of being able to extract important information from images, which reduces their sensitivity to changes in the direction and location of the camera with respect to the objects. \cite{ma2022vision}.

To train a deep neural network to transform perspective view (PV) images captured by an RGB camera into a bird's-eye-view (BEV) map, a large and diverse dataset is required to enhance the model's generalization and performance under various conditions. Collecting such data from real-world environments can be challenging and expensive, particularly as it necessitates obtaining corresponding top-down representations of the PV images. Therefore, we present a new approach for data collection tailored to our specific needs, providing a comprehensive dataset that can be utilized for similar research in the field of self-driving vehicles. Thus, the following can be summed up as our primary contributions:

\begin{itemize}
    \item Created a dataset using a simulator that produces PV images with vehicle bounding box location information, along with their corresponding BEV map which can also be applied to similar tasks in the field.
    \item Proposed an end-to-end deep learning architecture designed and tuned to transform PV images into accurate BEV maps.
\end{itemize}


\section{Related Work}

\subsection{Mapping from PV to BEV}

The task of transforming PV to BEV was mentioned in a couple of studies in the literature. A wide variety of approaches are employed to fulfill this objective. In the past, before the emergence and gigantic surge to DL, \cite{mallot1991inverse} tried to solve the problem by doing the mapping using a homography matrix. A computationally efficient and the earliest work that raises today's improvements in this field, yet having some flaws including the rigid flat-world assumption. Thanks to the recent enhancements in DL, researchers gain much more accuracy and efficiency in their methods, getting an extraordinary interest both from academics and industries. However, homography is not yet fully substituted by new methods and still are being used in recent studies \cite{gu2022homography}. 

Table \ref{tab:related} provides an overview of relevant studies relevant to this assignment. The main focus of our work is dedicated to those studies that empower DL techniques and generate 2D BEV maps of the ego vehicle's surrounding in which objects, especially other cars, are identified.

These studies can be divided into four groups based on the model proposed. In the first group, researchers harness a simple, yet effective, convolutional neural network (CNN) or propose a brand-new architecture to fulfill the objective. Given the fact that PV is completely located in a different space compared to the BEV, studies tried to tackle this variation problem using generative adversarial network (GAN), auto-encoder (AE), and empowering transformers along with CNNs, in the next three groups respectively to perform the transformation. Considering GANs, they can be considerably efficient in generating BEV maps corresponding to their ability to adversarially learn producing these maps, ensuring high accuracy by their discriminator. Related studies, also, change some features in this kind of Neural Network by adding an extra discriminator \cite{zhou2020pixel} or applying geometry \cite{zhu2018generative} as an additional attribute to the network, or proposing novel methods to adopt the difficulties and challenges presented in this task, resulting in a better performance. 

In the next group of studies, transformation is performed leveraging AE \cite{hoyer2019short}, variational AE (VAE) \cite{lu2019monocular}, and an ensemble of these models \cite{hendy2020fishing}. In this regard, given the characteristics of these architectures to encode the input images to a lower-dimensional latent space, providing a valuable set of information, they can decently generate corresponding BEV map representations. Within the final group, there are studies that leverage transformers (with cross-attention) to carry out the mapping objective by constructing queries and searching through the corresponding BEV image in order to gain beneficial features with an attention mechanism.

This is also worth mentioning that there are some studies that established multiple cameras around the ego vehicle to generate a 360$^{\circ}$ BEV map by combining features extracted in different angles \cite{reiher2020sim2real, gosala2022bird}, providing a high-value scene understanding of the ego vehicle's surroundings, relaxing the downstream tasks much more than before. The proposed method in this paper does not support this since we only installed a single camera parallel to the ground and facing frontward to capture PV images; However, this method is flexible and simple enough that can be easily modified to generate 360$^{\circ}$ BEV maps using previous techniques. Moreover, there are some studies that work with a sequence of images or videos instead of just a single image each time to take the best advantage of the information in previous or future frames, resulting in better accuracy and performance.

\begin{table*}[]
\centering
\caption{Related studies on BEV generation from PV.}
\begin{tabular}{llllrrl}
\hline
Ref.                                          & Model             & Modality & Accuracy                       & Year & \#Frames/Sequences & Dataset                     \\ \hline
\cite{palazzi2017learning}   & CNN (SDPN)              & image    & mIoU = 37.0\%, mDistance = 78m & 2017 & -                  & Synthesised (SVA)           \\
\cite{reiher2020sim2real}    & CNN               & image    & mIoU=80.9\%                    & 2020 & 33,000             & Synthesised                 \\ \hline
\cite{zhu2018generative}     & BridgeGAN         & image    & LPIPS=0.242                    & 2019 & 40,000             & SVA                         \\
\cite{fraser2021deepbev}     & Conditional GAN   & image    & mDistance=5.91m                & 2020 & 7,481              & KITTI, nuScenes             \\
\cite{zhou2020pixel}         & GAN               & image    & LPIPS=0.264                    & 2020 & -                  & SVA                         \\
\cite{jain2021generating}    & GAN               & sequence & RMSE=30.1                      & 2021 & -                  & SVA                         \\ \hline
\cite{lu2019monocular}       & VAE               & image    & mIoU=59.5\%                    & 2018 & 2,975              & Cityspcases, KITTI          \\
\cite{hoyer2019short}        & AE                & sequence & mIoU=0.61\%                    & 2019 & 59,500             & Cityspcases, KITTI          \\
\cite{hendy2020fishing}      & Ensemble of AE    & sequence & mIoU=44.3\%                    & 2020 & $\sim$2,000,000    & Synthesised, NuScenes, Lyft \\ \hline
\cite{roddick2020predicting} & CNN + Transformer & image    & mIoU=19.1\%                    & 2020 & -                  & NuScenes, Argoverse         \\
\cite{can2021structured}     & CNN + Transformer & image    & mIoU=14.9\%                    & 2021 & -                  & NuScenes                    \\
\cite{saha2022translating}   & CNN + Transformer & image    & mIoU=25.7\%                    & 2022 & -                  & NuScenes, Argoverse, Lyft   \\
\cite{gosala2022bird}        & CNN + Transformer & sequence & mIoU=33.65\%                   & 2022 & 702                & KITTI-360, NuScenes         \\ \hline
\end{tabular}
\label{tab:related}
\end{table*}

\subsection{Related Datasets}
In recent years, few datasets containing images from street spaces, pedestrians, and vehicles were publicly published mainly because of the ease of capturing RGB images and its low cost. Also, some of them provide other annotations such as bounding boxes for different objects and semantic segmented maps in various types of environmental and weather conditions as additional information. \cite{caesar2020nuscenes} is a popular dataset in this context which in addition to the RGB images that were captured using cameras placed in 6 different directions around the ego vehicle, provides Lidar and Radar images along with a 360$^{\circ}$ field of view. This dataset comprises 1000 scenes, Each of the 1000 scenes in this dataset has a duration of 20 seconds and is completely annotated with 3D bounding boxes for 23 classes and 8 attributes.  Also, \cite{geiger2013vision} is a similar dataset, providing Lidar images along with RGB images; However, it does not generate a 360$^{\circ}$ field of view and just contains PV images. In the next version of the dataset \cite{Liao2022PAMI}, 360$^{\circ}$ FoV is also added.

There are also some studies including \cite{palazzi2017learning}, that synthesized a brand-new dataset using a virtual city space simulator that replicates real-world objects and events, reducing the cost of acquisition, in this case, making the impossible task of taking BEV image along with corresponding PV image feasible. To the best of our knowledge, there is no specific real dataset that can directly address this task and researchers need to construct a pre-processing step before using the dataset \cite{gosala2022bird, jain2021generating}, increasing the chance of data loss and decelerate the process. Because DL models are data-hungry, they require a large volume of data in order to make more accurate predictions. In this study, we also proposed a huge dataset using a simulator to resolve the mentioned issue while generating more realistic images.


\section{Proposed Model}
We now introduce our suggested approach to the issue of creating a top-down representation of the environment for ego vehicles utilizing PV pictures and the locations of the bounding boxes of the vehicles within them. The fact that certain vehicles are partially unseen to the ego vehicle presents a problem while executing this transformation; as a result, it is not suitable to use classical or geometrical techniques in these situations due to the high probability of mistakes. Since a DL network can learn such situations and attempt to reduce the error, we can lessen their impact by applying DL techniques. 

Our model comprises two main sections, with the first section referred to as the input segment. This segment is responsible for gathering inputs, extracting their key characteristics, and converting them into a different feature space for subsequent utilization. The input segment is further divided into two branches. The initial branch employs a pre-trained backbone model to extract crucial features, such as spatial relative positions among objects in the input image. These features are then transformed into a feature vector, representing encoded attributes for further processing. Notably, the backbone model choice offers flexibility, enabling the utilization of different pre-trained models for the task at hand. Since the extracted features significantly influence the final outputs, employing state-of-the-art CNNs would enhance the overall system performance and accuracy. We elaborate on the impact of the backbone model's efficiency in the experimental section.

The second branch of our model is responsible for processing the location information of the bounding box of the selected vehicle in the front view image. This branch comprises multiple fully connected dense layers that encode these positions into another feature vector. Subsequently, these vectors are aggregated to establish the relationship between the features extracted from the first and second branches. This aggregation enables the model to locate selected vehicles in a top-down representation, thereby enhancing its ability in recognizing their spatial relations.

The second segment of our model comprises two branches dedicated to determining the location of the bounding box of the selected vehicle along the horizontal and vertical axes. This design allows the first branch to focus on the input image and its corresponding extracted features in the horizontal direction, specifically considering rows of the input image. Similarly, the second branch focuses on the vertical direction, taking into account the columns of the input image. To construct these branches, we utilize multiple fully connected dense layers, where the extracted features from the previous segment are fed into both branches.

Through our experimental investigations and statistical analysis of the dataset, we have observed that the majority of vehicles in the proposed dataset are vertically positioned. This observation aligns with real-world scenarios. Consequently, we have allocated more dense layers to the second branch of the second segment, which is responsible for generating the vertical locations of the bounding box.The decision was motivated by the notably greater standard deviation noted in the vertical value distribution as opposed to the horizontal value distribution.

\begin{figure*}[htbp]
\includegraphics[width=\textwidth]{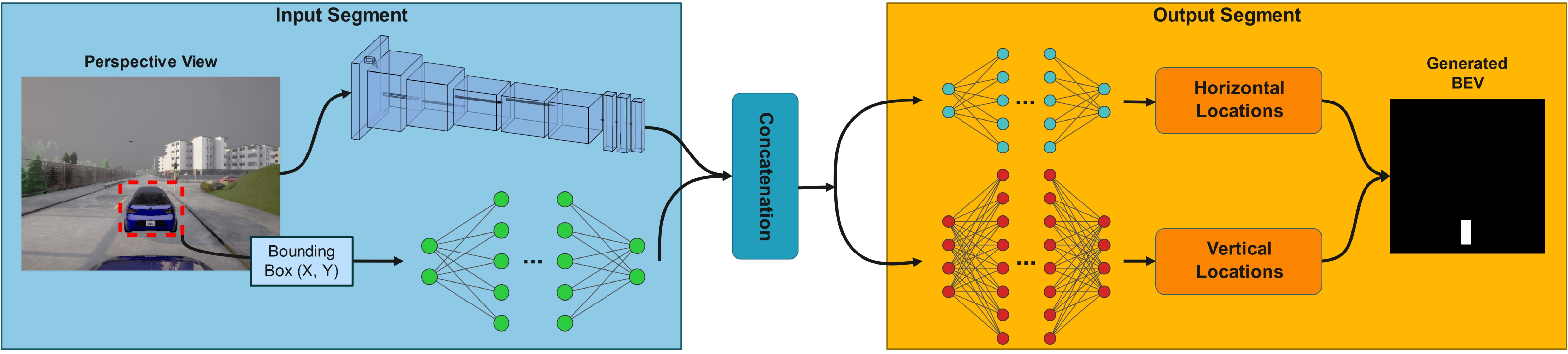}
\caption{The proposed architecture of deep neural network for predicting the bird-view.}
\label{fig_model}
\end{figure*}

Finally, the localization of the bounding box is accomplished by an aggregator script, which combines the generated horizontal and vertical locations to form the final bounding box. To optimize the model, we utilize the mean squared error (MSE) as our loss function. The MSE compares the predicted values in both the horizontal and vertical directions and, through the use of an optimizer, we aim to minimize this loss. In our final architecture, the ReLU activation function is employed for all layers, except for the output layers. The output layers utilize linear activation to facilitate the prediction of bounding boxes in pixel coordinates. A scheme of our proposed architecture is illustrated in Figure \ref{fig_model}.

\section{Proposed dataset}
The aforementioned challenges discussed in previous sections, including the lack of availability of a dataset that contains BEV maps along with corresponding PV images in real-world without any need for further pre-processing and interpretation, except for some synthesized datasets generated via simulation engines which were far away from realistic scenes, lead us to come up with new dataset, a more realistic one, providing a wide variety of different situation of street spaces, catalyzing researches in this field. We used an open-source simulation environment called CARLA \cite{dosovitskiy2017carla} which provides various city maps and efficient tools for simulating interactions of autonomous vehicles and developing different self-driving projects.
Moreover, we take advantage of \cite{datasetrunner, BEVgenerator} to develop our data generation pipeline, structure our code and stored data, and modify it to suit our task.

\subsection{Ego Vehicle Setup}
In order to prepare the dataset using the simulator, we spawned various ego vehicles in random spawning locations of the map as the origin of the output images, and we repeat this process in each epoch of data capturing so that the resulting dataset would have different records from divergent distributions and environments. Ego vehicle types are different in each epoch and it is chosen randomly from various options that the simulator offers. In each record in the final dataset, various valuable information about the ego car such as name, id, and speed is stored in a structured format.

\subsection{Sensors and synchronization}
We installed multiple sensors on the ego vehicle in order to provide various and valuable information in one go of the data acquisition process. First, an RGB camera, capturing PV images, is established parallel to the ground, facing frontward in the middle of the vehicle. Its attributes such as FOV, output image size, and relative location with respect to the ego vehicle were tuned in a way that produces PV images with desired aspect ratio synced with other sensors while covering the vehicle bonnet. Next, the depth camera sensor which provides a view over the scene codifying the distance of each pixel to the camera with the same attributes and location was installed; so the outputs of the RGB camera and depth sensor completely match each other. Some attributes of these sensors are described in detail in Table \ref{tab_sensor}.

Moreover, we establish a BEV generator pipeline, producing semantic segmented top-down representation respected to ego vehicle with synchronized aspect ratio with PV generated images, illustrated 8 various objects with different colors, is described further in Table \ref{tab_sensor}. Also, output maps are rotated to locate the ego car at the center and bottom of the BEV, which is always facing frontward, even when it is turning around. Finally, all objects in the simulated world are frizzed while data acquisition is proceeding in order to maintain the synchronization of the aforementioned output flows.






\begin{table}[]
\centering
\caption{Details and attributes associated with installed sensors and BEV generator pipeline.}
\begin{tabular}{@{}lp{7.6cm}@{}}
\toprule
Sensor & Details                                                                                                                                                                                                                                                                                      \\ \midrule
RGB    & 1024 × 768 resolution, 110$^{\circ}$ horizontal FOV, $\sim$800KB, iso = 100, gamma = 2.2, shutter speed = 200 per second, auto exposure, PNG                                                                                                                                                 \\ \\
Depth  & 1024 × 768 resolution, 110$^{\circ}$ horizontal FOV, $\sim$100KB, PNG, depth range = {[}0, 16777215{]}, normalized values                                                                                                                                                                    \\ \\
BEV    & 200 × 200 resolution, 4 pixels per meter ratio, covering 50 meters in horizontal and vertical aspects respected to ego vehicle, $\sim$10KB, segmented objects: ego vehicle, other vehicles, pedestrians, traffic lights and their state, roads, center lines, margin lines, navigation roads \\ \bottomrule
\end{tabular}
\label{tab_sensor}
\end{table}

\subsection{Maps}
We used 6 city maps, the one's simulator provides in its main release, 3D models of different towns, and their road definitions. these maps contain several types of roads (e.g. 2, 3, 4-lane roads), junctions, buildings, characteristics (e.g. urban, or rural), surface (e.g. flat or uneven), etc. Also, we define 5 different custom weather conditions based on daytime which are described in detail in Table \ref{tab_weather}. We also provide different climate categories such as rainy, sunny, and cloudy. These attributes make the generated dataset a generalized set, containing a wide variety of situations, giving us the ability to make our simulation as realistic as possible. The final dataset evenly contains images from these 6 maps.




\begin{table}[]
\centering
\caption{Various weather conditions implemented for the simulator.}
\resizebox{\columnwidth}{!}{%
\begin{tabular}{@{}llllr@{}}
\toprule
Type      & Cloudiness & Precipitation & Wind Intensity & Sun Altitude Angle \\ \midrule
Morning   & Low        & High          & Medium         & 30$^{\circ}$       \\
Midday    & Medium     & Low           & Medium         & 80$^{\circ}$       \\
Afternoon & High       & Low           & Medium         & 320$^{\circ}$      \\
Night     & Medium     & Medium        & Medium         & 300$^{\circ}$      \\
Default   & Low        & Low           & Low            & 0$^{\circ}$        \\ \bottomrule
\end{tabular}%
}
\label{tab_weather}
\end{table}

\subsection{Data annotation}
After collecting the raw data in each iteration of the data acquisition process from the aforementioned sensors, we annotate each vehicle bounding box both in PV and BEV by using the tools the simulator provides. Outputs only contain vehicles that have an equal or lower distance of 50 meters respected to the ego vehicle. Bounding boxes of vehicles were invisible vehicle PV, but are located at mentioned distance are also issued. The location of the top-left and bottom-right corners of each vehicle in PV and their corresponding locations in the BEV map were stored. Distances of other cars with respect to ego vehicles were calculated using the location of vehicles in the map of which simulator provided.

\subsection{Data acquisition}
The data acquisition process was repeated 6 times with 6 different maps. For each process, 100 vehicles were randomly selected from various available cars which the simulator offers and spawned in different locations. Each vehicle has an autopilot feature, providing a low-level intelligence for it to make the number of collisions as small as possible while making random decisions for navigation in the city. In each process, 20 different ego vehicles were selected consecutively. Each ego vehicle captured data for 20 continuous frames for 5 different weather conditions mentioned before. After a frame is stored, 5 frames will be skipped in order to prevent storing similar images, ensuring the generalization of situations with different possible combinations.

Finally, all the acquired data including images, maps, and extra information were stored in a structured HDF5 file, an open-source file format that supports large, complex, heterogeneous data that suits our task as well. An additional script is provided in order to convert stored records in this format into a raw and usable form for training the model. 
Table \ref{tab_dataset} provides statistical information about the proposed dataset, whereas Table \ref{tab_sample} presents a sample scheme of a single record in a dataset for each car spotted by ego vehicle.

\begin{figure}[htbp]
\centerline{\includegraphics[width=0.5\textwidth]{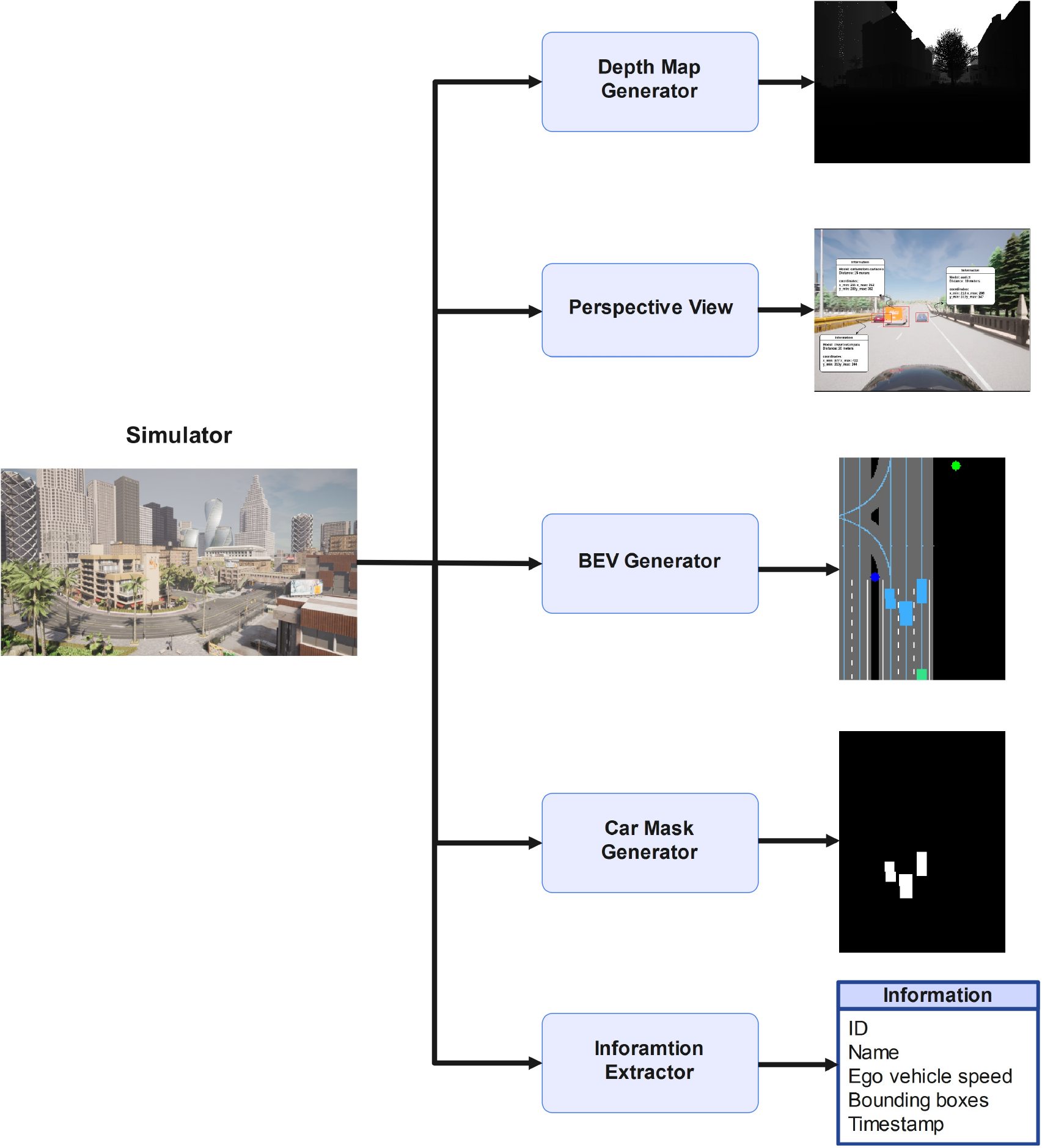}}
\caption{Data generating pipelines}
\label{fig_simulator}
\end{figure}


\begin{table}[]
\centering
\caption{statistical details of the proposed dataset.}
\begin{tabular}{@{}lr@{}}
\toprule
Attribute                             & Number \\ \midrule
Total number of images                & 12,000 \\
Total number of unique bounding boxes & 26,070 \\
FPS                                   & 30     \\
Size                                  & 108 GB \\
Generated vehicles in each map        & 100    \\
Different types of vehicles           & 26     \\ \bottomrule
\end{tabular}
\label{tab_dataset}
\end{table}

\begin{table}[]
\centering
\caption{Attributes in a single record in the proposed dataset.}
\begin{tabular}{@{}lp{5.9cm}@{}}
\toprule
Attribute          & Details                                                                                    \\ \midrule
ID                 & Identical number representing each vehicle in the map.                                     \\
Name               & Producer company along with its model.                                                     \\
Distance           & Distance from ego vehicle in meters.                                                       \\
Timestamp          & Time of issuing.                                                                           \\
RGB                & The PV output in the form of an RGB array captured by the front view camera.                  \\
Depth              & A normalized depth array captured by the depth sensor.                                     \\
Bounding box - RGB & Location of top-left and bottom-right pixels representing the selected vehicle in PV.      \\
Bounding box - BEV & Location of top-right and bottom-left pixels representing the selected vehicle in BEV map. \\
Speed              & Speed of ego vehicle during capturing the record.                                          \\ \bottomrule
\end{tabular}
\label{tab_sample}
\end{table}

\section{Experiments}

\subsection{Training protocol}

In order to assess our model, we divided our dataset into two separate and independent sets: a test set, which made up the remaining 10\% of the created dataset, and a training set, which made up 90\% of the total.
In terms of data augmentation, no specific techniques were required as our proposed dataset inherently provided a diverse range of input images with varying compositions. However, a normalization process was applied to the PV images to ensure consistent and standardized inputs throughout the training process.


Weights pre-trained on the ImageNet dataset were used to initialize the backbone models in the first branch of our model\cite{russakovsky2015imagenet}, while the remaining layers were randomly initialized. To optimize the network, we employed the Adam optimizer \cite{kingma2014adam} and trained the model for a total of 100 epochs on the training set. A batch size of 32 was used, along with a momentum of 0.9 and an initial learning rate of 0.001. During each training iteration, a single bounding box was selected from all the bounding boxes in a given PV image. This selected bounding box was then fitted into the model along with the corresponding PV image.

\subsection{Evaluations}


In order to assess the efficacy of our suggested methodology, we performed a comparison examination with prior research, implementing essential adjustments to align it with the framework of our dataset. Specifically, we trained the modified version of the prior work from scratch to ensure a fair comparison. We compared the performance of our model with two versions of SDPN, employing linear and tanh activation functions in the last layer. Furthermore, we evaluated the performance of our proposed model using various backbone architectures. The details of these backbone architectures are presented in Table \ref{tab_metrics}. For evaluation, we rely on four metrics, as described in \cite{palazzi2017learning}:

\paragraph{Intersection over Union (IoU)} The IoU metric quantifies the accuracy of the predicted bounding box $BB_p$ in relation to the target bounding box $BB_t$:

\begin{equation}
IoU(BB_p, BB_t) = \frac{Area(BB_p \cap BB_t)}{Area(BB_p \cup BB_t)}
\end{equation}

where $Area(BB)$ denotes the surface area of the bounding box $BB$.

\paragraph{Centroid Distance (CD)} CD is the Euclidean distance, measured in pixels, between the centers of the bounding boxes. It serves as an indicator of localization precision.\footnote{It's important to note that the target images have a resolution of 200 × 200, with each meter represented by 4 pixels.}

\paragraph{Height, Width Error (hE, wE)} hE and wE represent the average discrepancies in height and width of the bounding boxes respectively, expressed as a percentage relative to the ground truth dimensions.

\paragraph{Aspect Ratio Mean Error (arE)} arE measures the absolute difference in aspect ratio between the predicted and ground truth bounding boxes:

\begin{equation}
arE = \Bigg| \frac{BB_p \cdot width}{BB_p \cdot height} - \frac{BB_t \cdot width}{BB_t \cdot height} \Bigg|
\end{equation}

The results of our evaluation indicate that our proposed model demonstrates superior performance compared to the previous study across various metrics and configurations. Notably, when considering each metric and configuration, our model consistently outperformed the previous study. Specifically, we observed that the model utilizing a ResNet-50 \cite{he2016deep} backbone achieved more precise predictions of the height, width, and aspect ratio of bounding boxes compared to other configurations. This highlights the significant impact of the choice of backbone architecture on the predictive capabilities of our proposed method.

Alternatively, VGG \cite{simonyan2014very} backbone models exhibited greater accuracy in predicting the location of the center of bounding boxes compared to ResNet backbone models. However, it is important to note that the difference in centroid distances between the two was not significantly high. Nonetheless, the proposed model with ResNet backbone demonstrated higher IoU values, indicating better overall performance. These findings emphasize the influence of the backbone architecture on the prediction accuracy of our proposed method. Specifically, the ResNet backbone architecture yields superior performance in terms of predicting bounding box characteristics, resulting in higher IoU values.




\begin{table}[]
\centering
\caption{Results of the performance of the proposed model against the baseline.}
\begin{tabular}{@{}lrrrrr@{}}
\toprule
Model                   & IoU             & CD              & hE              & wE              & arE             \\ \midrule
SDPN-tanh               & 0.1582          & -               & 2.8726          & 1.2225          & 3.8878          \\
SDPN-linear             & 0.1318          & 4.3232          & 0.9657          & 0.5134          & 1.6710         \\ \midrule
ProposedMethod-VGG16    & 0.1190         & \textbf{3.5033} & 0.0571          & 0.1045          & 0.4298          \\
ProposedMethod-VGG19    & 0.0137         & 4.9355          & 0.2974          & 0.1356          & 0.7286          \\
ProposedMethod-Resnet50 & \textbf{0.1897} & 4.4275          & \textbf{0.0089} & \textbf{0.0744} & \textbf{0.5506} \\ \bottomrule
\end{tabular}
\label{tab_metrics}
\end{table}

\section{Conclusion}

In this paper, we propose an end-to-end deep learning architecture that generates Bird's Eye View (BEV) maps from PV images and bounding box location information. We introduce a novel data generator pipeline that provides PV images and corresponding BEV maps, eliminating the need for extensive pre-processing. Through evaluations of our dataset, we demonstrate that our model outperforms existing baselines. Future work includes constructing 3D BEV maps, incorporating transformers for improved mapping, expanding the dataset, and integrating multiple cameras for a comprehensive view. Our approach contributes to advancing research in this field by providing a realistic dataset and showcasing the effectiveness of our model in generating accurate BEV maps.

\bibliographystyle{IEEEtran}
\bibliography{ref}

\end{document}